\documentclass[runningheads]{llncs}
\usepackage[T1]{fontenc}
\usepackage{multirow}
\usepackage{graphicx,verbatim}
\usepackage{booktabs}
\usepackage{xurl}
\usepackage{hyperref}
\usepackage{graphicx}
\usepackage{subcaption}
\usepackage{amsmath}
\usepackage[dvipsnames]{xcolor}
\usepackage{todonotes}
\usepackage[misc]{ifsym}
\begin{document}
\title{Representation Learning for 3D Brain Imaging: A Benchmark}

\author{Pierre Falconnier\inst{1}\orcidID{0009-0002-8281-3568} (\text{Email}) \and
 Robin Trombetta\inst{1}\orcidID{0009-0001-7945-6953} \and
Nicolas Duchateau\inst{1,2}\orcidID{0000-0001-8803-2004} \and
Carole Lartizien\inst{1}\orcidID{0000-0001-7594-4231}}

\authorrunning{P. Falconnier et al.}
%
\institute{Univ. Lyon, CNRS UMR 5220, Inserm U1294, INSA Lyon, UCBL, CREATIS, France \and Institut Universitaire de France (IUF) \\
\email{pierre.falconnier@creatis.insa-lyon.fr}}



\maketitle              

\begin{abstract}

Many deep learning architectures for computer vision tasks rely on features extracted by a pre-trained encoder on large databases of natural 2D RGB images. However, the wide domain gap with medical images hampers the direct transfer of optimized deep learning models from computer vision to clinical applications. Recently, self-supervised pre-training and foundation models emerged as promising alternatives to mitigate these limitations. Yet, constructing such models for medical imaging faces critical challenges, including designing and adapting training strategies for large 3D volume datasets, and the need for standardized benchmarks to evaluate feature extraction performance. In this study, we propose an extensive comparison of vision encoders with a focus on 3D brain MRI. First, we consider 8 fully and self-supervised models pre-trained on natural images (e.g. ResNet, DINO, CLIP) and 10 backbone architectures trained on medical images, either based on supervised (e.g. RadImageNet) or self-supervised (e.g. BiomedCLIP, SimCLR) tasks. All pre-trained encoders are evaluated on various downstream tasks to assess the representation power of their latent representations, including age and sex prediction as well as disease classification on the ADNI and PPMI datasets. Our study shows that 2D encoders pre-trained on natural images perform the most favorably, even compared to models pre-trained on more than 100k brain MRI volumes, highlighting the need for more suitable and robust methods for representation learning in 3D medical imaging. The source code for this study is publicly available at: \url{https://github.com/PierreFalconnier/brain-representation-benchmark}.


\end{abstract}

\section{Introduction}

Deep learning relies on the automatic learning of representations from images. 
In generic computer vision (CV), where annotated data is more abundant, a major research focus has been on learning robust representations from images. Since the emergence of pioneering models like ResNet \cite{resnet}, numerous advancements have been made, including modern architectures such as Vision Transformers (ViT) \cite{vit}. At the same time, the self-supervised learning (SSL) paradigm has emerged and quickly gained in popularity. Built on the conception of relevant pretext tasks that do not require annotations, SSL enables learning representations that are often more generic and robust, agnostic to any downstream task. SSL paved the way for a new paradigm: foundation models, with successful methods such as SimCLR \cite{simclr}, Masked Auto-Encoder (MAE) \cite{mae}, or DINOv2 \cite{dinov2}. These models have reshaped the landscape of automatic image processing by enabling the extraction of powerful features and showcasing great zero- and few-shot performance. Many neural networks architectures make use of such models pretrained on large natural image dataset (>10M images), e.g. the LVD-142M dataset,  as frozen backbones to extract meaningful representations from input images \cite{izquierdo2024dino_vpr,damm2025anomalydino}.\\

In medical imaging (MI), scarcity of data and limited availability of full or weak annotations affect the robustness of models, which may overfit representations within their training domain with performance degrading when evaluated on external data. Scientific studies are partially shifting toward training increasingly large models using ever larger amounts of data. Following the advent of foundation models and general-purpose models in CV, a lot of recent works have developed polyvalent networks for MI analysis \cite{Wasserthal2023total_segmentator}. SSL has also been increasingly explored, primary as a pre-training step to enhance performance in specific downstream tasks \cite{manna2024selfsupervised}. 
In particular for neuroimaging, thanks to highly valuable worldwide efforts towards data sharing, SSL has recently moved to an unprecedented scale. OpenMind \cite{wald2025openmind} is an open database of 114k multi-modal 3D MR head and neck images of healthy and pathological subjects that has been used to benchmark standard SSL methods. Similarly, the FOMO25 challenge \cite{munk2026fomo25}
evaluated 
19 foundation models 
for representation learning in neuroimaging based on dataset of 60k multimodal brain MRI, with an ongoing extension toward more than 300k samples.

\begin{figure}[t]
\centering
\includegraphics[width=\textwidth]{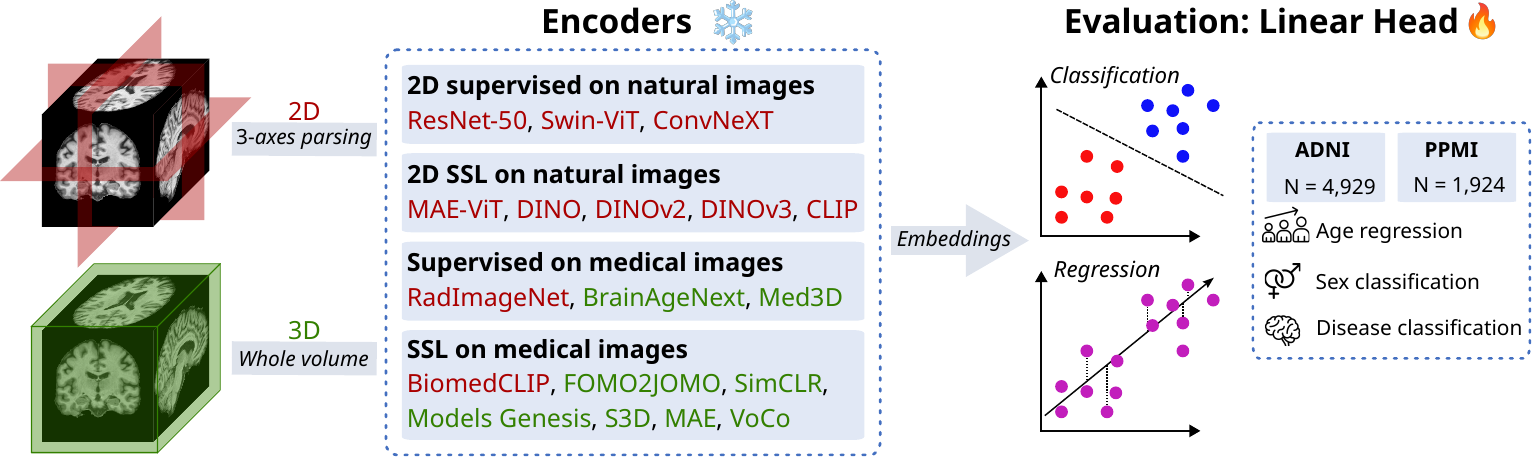}
\caption{Overview of the proposed benchmarking framework.} 
\label{fig1:schema}
\end{figure}

In this work, we propose 
an extensive benchmark of feature extraction methods, of high relevance given the growing number of representation learning methods and the increasing need for extracting robust, annotation-efficient, and domain-specific features in the medical imaging field.
We focus on 3D brain MR imaging and 
rigorously compare 
18 pre-trained encoders
taken from the field of generic CV or specialized for MI analysis.
Contrary to previous related works \cite{huang2023selfsupervised,zhang2023divessl}, our benchmark analysis focuses on assessing the representation power 
of the encoders directly at the feature level by computing the performance of the learned and frozen representations for different downstream tasks without full fine-tuning. 
Our work shows that despite having been trained on a 
large 
number of MR volumes, most recent SSL foundation models for neuroimaging are still surpassed by 2D encoders derived from general CV.
Our study is highly relevant for 
the development of deep learning algorithms for medical image analysis by setting a fair, extensive and reliable evaluation of state-of-the-art representation learning methods and pushing towards the design of new representation learning models dedicated to 3D brain medical imaging.

\section{Methodology} \label{methodo}

\subsection{Overview}

Our framework consists of three key components (Fig.~\ref{fig1:schema}).
We select an exhaustive series of image encoding models including frozen pre-trained models from the CV and MI domains. 
Then, we perform inference using a adaptation strategy for 2D-based models, while 3D models are applied directly, to extract feature representations from 3D MRI scans. 
Finally, we evaluate the representation ability of the features by assessing their performance on different downstream tasks.

\subsection{Pre-trained encoders}
\label{subsec:pretrained_encoders}

We selected 18 pre-trained encoders from the literature. 
We included standard general-purpose backbones pre-trained in a supervised fashion on ImageNet, widespread SSL foundation models for image processing as well as models pre-trained on medical images. \\
\textbf{Supervised encoders on natural images.} We included the standard ResNet-50 \cite{resnet}, the Vision Transformer with shifted windows called Swin-ViT \cite{swinvit}, the CNN ConvNeXt \cite{convnext}. All the models are trained solely on ImageNet with supervised loss. Swint-ViT has been preferred to classical ViT after internal tests. \\ 
\textbf{SSL models on natural images}. We included encoders trained on natural images with a self-supervised loss. In this category, our benchmark incorporates MAE-ViT, a Vision Transformer trained with the Masked Auto-Encoder paradigm \cite{mae}, the three iterations of the model DINO, namely DINO \cite{dino}, DINOv2 \cite{dinov2} and DINOv3 \cite{dinov3}. Finally, we 
included the ResNet-50 encoder from CLIP \cite{clip}, a SSL foundation model trained with paired image and text. \\
\textbf{Supervised models for MI.} 
We included the ResNet50 architecture trained on RadImageNet \cite{2022radimagenet}, an aggregation of 
more than 1.3M 
2D medical images of different modalities (MRI, CT, US) and organs (brain, chest, etc.). We also included BrainAgeNext, a CNN trained to predict the biological age of subjects from their T1w 3D MRI, which 
ranked second at the Open Track of FOMO25 Challenge. Our benchmark also contains Med3D \cite{chen2019med3d}, a general 3D feature extractor trained on diverse body regions and imaging modalities (MRI and CT). Other popular models in this category
were not considered either because they were not trained on datasets including MR images or because their weights were not available. \\
\textbf{SSL models for MI.} This category includes BiomedCLIP \cite{zhang2023biomedclip}, an adaptation of CLIP to medical image–text pairs, the FOMO2JOMO model, which won the method track of the FOMO25 challenge, and five models trained with different SSL methods on the OpenMind dataset \cite{wald2025openmind}: SimCLR \cite{simclr}, Models Genesis \cite{zhou2021models_genesis}, S3D \cite{wald2025s3d}, MAE \cite{mae} and VoCo \cite{wu2024voco}. \\
\textbf{Adaptation of 2D encoders}. Among all previously presented backbones,
10 (written in red in Fig.~\ref{fig1:schema})  out of 18 are designed to take 2D images as inputs and thus need adaptation to extract features from 3D brain volumes (see Fig. \ref{fig1:schema}). To do so, we parse the 3D volume slice-wise along each of the three axes, extract a feature vector for each 2D slice, apply average pooling to the features of each axis separately, and concatenate the three resulting vectors to summarize the 3D image content. Other possible ways to encode 3D volumes with 2D encoders could have been single-axis slice parsing, or more advanced techniques, such as ACS convolutions \cite{yang2021acsconv} or Raptor \cite{an2025raptor}, but internal tests indicated that our proposed multi-axis slice-wise feature extraction performs best. 

\section{Experiments}

\subsection{Data and preprocessing}

Our experiments were conducted on the following datasets:

\href{https://adni.loni.usc.edu/}{\textbf{ADNI}}. 3D T1w images from 1,674 cognitively normal (CN) and Alzheimer (AD) subjects were collected from the ADNI database, corresponding to a total of 4,929 visits (3,763 CN and 1,166 AD visits). For a given visit, only the first acquisition was retained. For the age regression task, only cognitively normal subjects were included. The sex balance is 52.3\%F, and the mean age on the CN subset is $74.7 \pm 7.4$ years.\

\href{https://www.ppmi-info.org/}{\textbf{PPMI}}. 3D T1w images from 1,061 cognitively normal (CN) and Parkinson (PD) subjects subjects were collected from the PPMI database, corresponding to 1,924 visits (266 CN and 1,658 PD visits). For a given visit, only the first acquisition was retained. For the age regression task, only healthy controls were included. The sex balance is 36.5\%F, and the mean age on the CN subset is  $61.4 \pm 11.8$ years.

These two datasets are continuously updated. The version used in this study was accessed in late February 2026. We verified that there was no overlap between these two datasets and the datasets used to train the models described in Section \ref{subsec:pretrained_encoders}. All images of both ADNI and PPMI datasets were pre-processed with Turboprep\footnote{\href{https://github.com/LemuelPuglisi/turboprep}{https://github.com/LemuelPuglisi/turboprep}}, which includes standard pre-processing steps for 3D brain MRI, namely N4 bias field correction, skull stripping, and affine registration to MNI-152 atlas. Low-quality results were excluded. The intensities of the volumes were clipped based on the 0.5th and 99.5 percentiles and scaled to $[0,1]$. The volumes were cropped or padded to $160 \times 192 \times 160$ voxels. Finally, for each pre-trained network, the input slices or volumes were further processed to satisfy the shape and intensity range required by each model (e.g. repeat channels, standardize, etc.).

\subsection{Evaluation procedure}

To assess the expressiveness of the latent space of all 18 backbones, three main tasks were considered for each dataset (Fig. \ref{fig1:schema}). The first task is sex classification with linear probing. The second one is age linear regression, performed exclusively on healthy subjects. The third task is the diagnosis with linear probing of Alzheimer's or Parkinson's disease. For all models and tasks, the encoders weights were frozen and the linear heads of the classification or regression models were fitted to the output dimension of the extracted latent representation vectors. 
For each database and task, 20\% of the data were used to tune the hyperparameters of the linear classifier or regressor, and the remaining 80\% were used for the final evaluation. The hyperparameter search was performed using 5-fold stratified cross-validation and involved determining the optimal regularization strength, which ranged from $10^{-6}$ to $10^{4}$, as well as testing a setting without regularization. The final performance of the models was evaluated using 10,000 bootstrap iterations with the 0.632 method to compute the mean metric values and their 90\% confidence intervals. The evaluation metrics are balanced accuracy for classification tasks and mean absolute error for the regression task. Care was taken to ensure that no subject-level or visit-level data leakage occurred across the initial split, cross-validation folds or during the bootstrap procedure.
The same bootstrap resampling indices were used across all models. 
For each pair of models, we computed the sign of paired differences between their scores on each resample and derived a two-sided p-value.

\section{Results}

\begin{table}[h!]
\centering
\footnotesize

\caption{Results of all the compared models on the six metrics of interest (balanced accuracy and mean absolute error for classification and regression tasks respectively). We report the mean and the 90\% confidence interval, computed using bootstrap sampling. For each metric, the best performing model is highlighted in \textbf{bold} and the second-best is \underline{underlined}. The last column shows the (rounded) median rank obtained by each model across the six tasks. 
The symbol $^{\dagger}$ indicates the models for which the paired bootstrap test with the top-performing model is significant at the level $p=0.05$.}
\label{tab:results}
\resizebox{\linewidth}{!}{%
\begin{tabular}{@{}lcc@{\hspace{4pt}}cccc@{\hspace{4pt}}c@{}}
\toprule
\multirow{2}{*}{\textbf{Models}} & \multicolumn{3}{c}{\textbf{PPMI}} & \multicolumn{3}{c}{\textbf{ADNI}} & \multirow{3}{*}{\begin{tabular}[c]{@{}c@{}}\textbf{Median} \\ \textbf{rank}\end{tabular}} \\ 
\cmidrule(lr){2-4} \cmidrule(lr){5-7}
& Sex $\uparrow$ & Age $\downarrow$ & CN/PD $\uparrow$  & \phantom{} Sex $\uparrow$ & Age $\downarrow$ & CN/AD $\uparrow$ &   \\ \midrule
ResNet50 (R50)
    & \begin{tabular}[c]{@{}c@{}}0.916\\ {[}0.895--0.935{]}\end{tabular}
    & \begin{tabular}[c]{@{}c@{}}4.47\\ {[}3.63--6.49{]}\end{tabular}
    & \begin{tabular}[c]{@{}c@{}}0.596\\ {[}0.559--0.633{]}\end{tabular}
    & \begin{tabular}[c]{@{}c@{}}0.931\\ {[}0.916--0.946{]}\end{tabular}
    & \begin{tabular}[c]{@{}c@{}}\textbf{3.00}\\ {[}2.84--3.16{]}\end{tabular}
    & \begin{tabular}[c]{@{}c@{}}0.838\\ {[}0.817--0.860{]}\end{tabular}
    & 5 \\[6pt]
R50-CLIP
    & \begin{tabular}[c]{@{}c@{}}0.925\\ {[}0.905--0.944{]}\end{tabular}
    & \begin{tabular}[c]{@{}c@{}}4.49\\ {[}3.83--5.16{]}\end{tabular}
    & \begin{tabular}[c]{@{}c@{}}0.607\\ {[}0.568--0.645{]}\end{tabular}
    & \begin{tabular}[c]{@{}c@{}}0.922\\ {[}0.907--0.937{]}\end{tabular}
    & \begin{tabular}[c]{@{}c@{}}3.52$^{\dagger}$\\ {[}3.38--3.67{]}\end{tabular}
    & \begin{tabular}[c]{@{}c@{}}0.779$^{\dagger}$\\ {[}0.757--0.799{]}\end{tabular}
    & 8 \\[6pt]
ConvNeXt
    & \begin{tabular}[c]{@{}c@{}}\underline{0.937}\\ {[}0.917--0.955{]}\end{tabular}
    & \begin{tabular}[c]{@{}c@{}}4.24\\ {[}3.59--4.90{]}\end{tabular}
    & \begin{tabular}[c]{@{}c@{}}0.599\\ {[}0.567--0.632{]}\end{tabular}
    & \begin{tabular}[c]{@{}c@{}}0.936\\ {[}0.921--0.950{]}\end{tabular}
    & \begin{tabular}[c]{@{}c@{}}3.47$^{\dagger}$\\ {[}3.33--3.62{]}\end{tabular}
    & \begin{tabular}[c]{@{}c@{}}0.830$^{\dagger}$\\ {[}0.807--0.852{]}\end{tabular}
    & \underline{3} \\[6pt]
SwinViT
    & \begin{tabular}[c]{@{}c@{}}0.937\\ {[}0.918--0.955{]}\end{tabular}
    & \begin{tabular}[c]{@{}c@{}}4.75\\ {[}4.06--5.47{]}\end{tabular}
    & \begin{tabular}[c]{@{}c@{}}0.599\\ {[}0.569--0.631{]}\end{tabular}
    & \begin{tabular}[c]{@{}c@{}}0.935\\ {[}0.920--0.949{]}\end{tabular}
    & \begin{tabular}[c]{@{}c@{}}3.53$^{\dagger}$\\ {[}3.39--3.67{]}\end{tabular}
    & \begin{tabular}[c]{@{}c@{}}0.813$^{\dagger}$\\ {[}0.792--0.834{]}\end{tabular}
    & 7 \\[6pt] \midrule
ViT-DINO
    & \begin{tabular}[c]{@{}c@{}} 0.929\\ {[}0.909--0.947{]}\end{tabular}
    & \begin{tabular}[c]{@{}c@{}}4.56\\ {[}3.95--5.19{]}\end{tabular}
    & \begin{tabular}[c]{@{}c@{}}0.613\\ {[}0.576--0.649{]}\end{tabular}
    & \begin{tabular}[c]{@{}c@{}}\underline{0.940}\\ {[}0.925--0.954{]}\end{tabular}
    & \begin{tabular}[c]{@{}c@{}}\underline{3.13}\\ {[}2.98--3.28{]}\end{tabular}
    & \begin{tabular}[c]{@{}c@{}}\underline{0.852}\\ {[}0.830--0.873{]}\end{tabular}
    & 4 \\[6pt]
ViT-DINOv2
    & \begin{tabular}[c]{@{}c@{}}\textbf{0.938}\\ {[}0.920--0.955{]}\end{tabular}
    & \begin{tabular}[c]{@{}c@{}}\textbf{4.11}\\ {[}3.50--4.77{]}\end{tabular}
    & \begin{tabular}[c]{@{}c@{}}0.598\\ {[}0.567--0.628{]}\end{tabular}
    & \begin{tabular}[c]{@{}c@{}}\textbf{0.948}\\ {[}0.935--0.961{]}\end{tabular}
    & \begin{tabular}[c]{@{}c@{}}3.21$^{\dagger}$\\ {[}3.06--3.35{]}\end{tabular}
    & \begin{tabular}[c]{@{}c@{}}0.827$^{\dagger}$\\ {[}0.804--0.850{]}\end{tabular}
    & \textbf{2} \\[6pt]
ViT-DINOv3
    & \begin{tabular}[c]{@{}c@{}}0.932\\ {[}0.913--0.950{]}\end{tabular}
    & \begin{tabular}[c]{@{}c@{}}4.62\\ {[}3.96--5.29{]}\end{tabular}
    & \begin{tabular}[c]{@{}c@{}}0.596\\ {[}0.561--0.630{]}\end{tabular}
    & \begin{tabular}[c]{@{}c@{}}0.934\\ {[}0.919--0.948{]}\end{tabular}
    & \begin{tabular}[c]{@{}c@{}}3.24$^{\dagger}$\\ {[}3.10--3.38{]}\end{tabular}
    & \begin{tabular}[c]{@{}c@{}}0.822$^{\dagger}$\\ {[}0.801--0.842{]}\end{tabular}
    & 6 \\[6pt]
ViT-MAE
    & \begin{tabular}[c]{@{}c@{}}0.932\\ {[}0.912--0.950{]}\end{tabular}
    & \begin{tabular}[c]{@{}c@{}}4.94$^{\dagger}$\\ {[}4.25--5.67{]}\end{tabular}
    & \begin{tabular}[c]{@{}c@{}}0.606\\ {[}0.566--0.645{]}\end{tabular}
    & \begin{tabular}[c]{@{}c@{}}0.935\\ {[}0.921--0.949{]}\end{tabular}
    & \begin{tabular}[c]{@{}c@{}}3.57$^{\dagger}$\\ {[}3.43--3.71{]}\end{tabular}
    & \begin{tabular}[c]{@{}c@{}}0.809$^{\dagger}$\\ {[}0.788--0.829{]}\end{tabular}
    & 8 \\ \midrule \midrule
BrainAgeNext
    & \begin{tabular}[c]{@{}c@{}}0.645$^{\dagger}$\\ {[}0.615--0.676{]}\end{tabular}
    & \begin{tabular}[c]{@{}c@{}}7.84$^{\dagger}$\\ {[}7.04--8.64{]}\end{tabular}
    & \begin{tabular}[c]{@{}c@{}}0.580\\ {[}0.540--0.619{]}\end{tabular}
    & \begin{tabular}[c]{@{}c@{}}0.699$^{\dagger}$\\ {[}0.678--0.718{]}\end{tabular}
    & \begin{tabular}[c]{@{}c@{}}4.59$^{\dagger}$\\ {[}4.43--4.75{]}\end{tabular}
    & \begin{tabular}[c]{@{}c@{}}0.713$^{\dagger}$\\ {[}0.691--0.734{]}\end{tabular}
    & 13 \\[6pt]
RadImageNet
    & \begin{tabular}[c]{@{}c@{}}0.921\\ {[}0.900--0.941{]}\end{tabular}
    & \begin{tabular}[c]{@{}c@{}}\underline{4.16}\\ {[}3.53--4.81{]}\end{tabular}
    & \begin{tabular}[c]{@{}c@{}}0.594\\ {[}0.559--0.630{]}\end{tabular}
    & \begin{tabular}[c]{@{}c@{}}0.923\\ {[}0.906--0.938{]}\end{tabular}
    & \begin{tabular}[c]{@{}c@{}}3.15\\ {[}2.97--3.33{]}\end{tabular}
    & \begin{tabular}[c]{@{}c@{}}\textbf{0.857}\\ {[}0.836--0.878{]}\end{tabular}
    & 5 \\[6pt]
R50-Med3D
    & \begin{tabular}[c]{@{}c@{}}0.738$^{\dagger}$\\ {[}0.712--0.763{]}\end{tabular}
    & \begin{tabular}[c]{@{}c@{}}7.56$^{\dagger}$\\ {[}6.83--8.35{]}\end{tabular}
    & \begin{tabular}[c]{@{}c@{}}0.615\\ {[}0.578--0.652{]}\end{tabular}
    & \begin{tabular}[c]{@{}c@{}}0.709$^{\dagger}$\\ {[}0.687--0.731{]}\end{tabular}
    & \begin{tabular}[c]{@{}c@{}}4.57$^{\dagger}$\\ {[}4.39--4.76{]}\end{tabular}
    & \begin{tabular}[c]{@{}c@{}}0.750$^{\dagger}$\\ {[}0.728--0.771{]}\end{tabular}
    & 12 \\[6pt] \midrule
BiomedCLIP
    & \begin{tabular}[c]{@{}c@{}}0.892$^{\dagger}$\\ {[}0.869--0.914{]}\end{tabular}
    & \begin{tabular}[c]{@{}c@{}}4.79\\ {[}4.23--5.39{]}\end{tabular}
    & \begin{tabular}[c]{@{}c@{}}\textbf{0.624}\\ {[}0.587--0.659{]}\end{tabular}
    & \begin{tabular}[c]{@{}c@{}}0.893$^{\dagger}$\\ {[}0.877--0.907{]}\end{tabular}
    & \begin{tabular}[c]{@{}c@{}}3.64$^{\dagger}$\\ {[}3.51--3.79{]}\end{tabular}
    & \begin{tabular}[c]{@{}c@{}}0.806$^{\dagger}$\\ {[}0.785--0.826{]}\end{tabular}
    & 9 \\[6pt]
FOMO2JOMO
    & \begin{tabular}[c]{@{}c@{}}0.798$^{\dagger}$\\ {[}0.775--0.821{]}\end{tabular}
    & \begin{tabular}[c]{@{}c@{}}7.28$^{\dagger}$\\ {[}6.54--8.04{]}\end{tabular}
    & \begin{tabular}[c]{@{}c@{}}0.572\\ {[}0.526--0.620{]}\end{tabular}
    & \begin{tabular}[c]{@{}c@{}}0.806$^{\dagger}$\\ {[}0.790--0.821{]}\end{tabular}
    & \begin{tabular}[c]{@{}c@{}}4.14$^{\dagger}$\\ {[}4.00--4.28{]}\end{tabular}
    & \begin{tabular}[c]{@{}c@{}}0.777$^{\dagger}$\\ {[}0.759--0.796{]}\end{tabular}
    & 11 \\[6pt]
R50-OM-MAE
    & \begin{tabular}[c]{@{}c@{}}0.880$^{\dagger}$\\ {[}0.856--0.904{]}\end{tabular}
    & \begin{tabular}[c]{@{}c@{}}5.31$^{\dagger}$\\ {[}4.71--5.94{]}\end{tabular}
    & \begin{tabular}[c]{@{}c@{}}\underline{0.574}\\ {[}0.529--0.618{]}\end{tabular}
    & \begin{tabular}[c]{@{}c@{}}0.854$^{\dagger}$\\ {[}0.838--0.870{]}\end{tabular}
    & \begin{tabular}[c]{@{}c@{}}3.82$^{\dagger}$\\ {[}3.67--3.97{]}\end{tabular}
    & \begin{tabular}[c]{@{}c@{}}0.797$^{\dagger}$\\ {[}0.776--0.817{]}\end{tabular}
    & 14 \\[6pt]
R50-OM-MG
    & \begin{tabular}[c]{@{}c@{}}0.851$^{\dagger}$\\ {[}0.825--0.875{]}\end{tabular}
    & \begin{tabular}[c]{@{}c@{}}5.82$^{\dagger}$\\ {[}5.09--6.60{]}\end{tabular}
    & \begin{tabular}[c]{@{}c@{}}0.610\\ {[}0.571--0.648{]}\end{tabular}
    & \begin{tabular}[c]{@{}c@{}}0.855$^{\dagger}$\\ {[}0.838--0.871{]}\end{tabular}
    & \begin{tabular}[c]{@{}c@{}}3.94$^{\dagger}$\\ {[}3.79--4.09{]}\end{tabular}
    & \begin{tabular}[c]{@{}c@{}}0.794$^{\dagger}$\\ {[}0.772--0.816{]}\end{tabular}
    & 16 \\[6pt]
R50-OM-S3D
    & \begin{tabular}[c]{@{}c@{}}0.794$^{\dagger}$\\ {[}0.768--0.820{]}\end{tabular}
    & \begin{tabular}[c]{@{}c@{}}6.25$^{\dagger}$\\ {[}5.45--7.10{]}\end{tabular}
    & \begin{tabular}[c]{@{}c@{}}0.580\\ {[}0.539--0.620{]}\end{tabular}
    & \begin{tabular}[c]{@{}c@{}}0.859$^{\dagger}$\\ {[}0.842--0.875{]}\end{tabular}
    & \begin{tabular}[c]{@{}c@{}}3.86$^{\dagger}$\\ {[}3.72--4.00{]}\end{tabular}
    & \begin{tabular}[c]{@{}c@{}}0.808$^{\dagger}$\\ {[}0.789--0.828{]}\end{tabular}
    & 14 \\[6pt]
R50-OM-SimCLR
    & \begin{tabular}[c]{@{}c@{}}0.879$^{\dagger}$\\ {[}0.855--0.901{]}\end{tabular}
    & \begin{tabular}[c]{@{}c@{}}4.98$^{\dagger}$\\ {[}4.39--5.62{]}\end{tabular}
    & \begin{tabular}[c]{@{}c@{}}0.608\\ {[}0.570--0.647{]}\end{tabular}
    & \begin{tabular}[c]{@{}c@{}}0.871$^{\dagger}$\\ {[}0.852--0.889{]}\end{tabular}
    & \begin{tabular}[c]{@{}c@{}}3.57$^{\dagger}$\\ {[}3.43--3.71{]}\end{tabular}
    & \begin{tabular}[c]{@{}c@{}}0.850\\ {[}0.831--0.868{]}\end{tabular}
    & 16 \\[6pt]
R50-OM-VoCo
    & \begin{tabular}[c]{@{}c@{}}0.798$^{\dagger}$\\ {[}0.772--0.824{]}\end{tabular}
    & \begin{tabular}[c]{@{}c@{}}5.92$^{\dagger}$\\ {[}5.24--6.62{]}\end{tabular}
    & \begin{tabular}[c]{@{}c@{}}0.619\\ {[}0.586--0.656{]}\end{tabular}
    & \begin{tabular}[c]{@{}c@{}}0.777$^{\dagger}$\\ {[}0.757--0.797{]}\end{tabular}
    & \begin{tabular}[c]{@{}c@{}}4.17$^{\dagger}$\\ {[}4.01--4.34{]}\end{tabular}
    & \begin{tabular}[c]{@{}c@{}}0.752$^{\dagger}$\\ {[}0.729--0.774{]}\end{tabular}
    & 16 \\[6pt] 
\bottomrule
\end{tabular}
}
\end{table}

\begin{figure}[t]
     \centering
     \begin{subfigure}{0.49\textwidth}
         \centering
         \includegraphics[width=\linewidth]{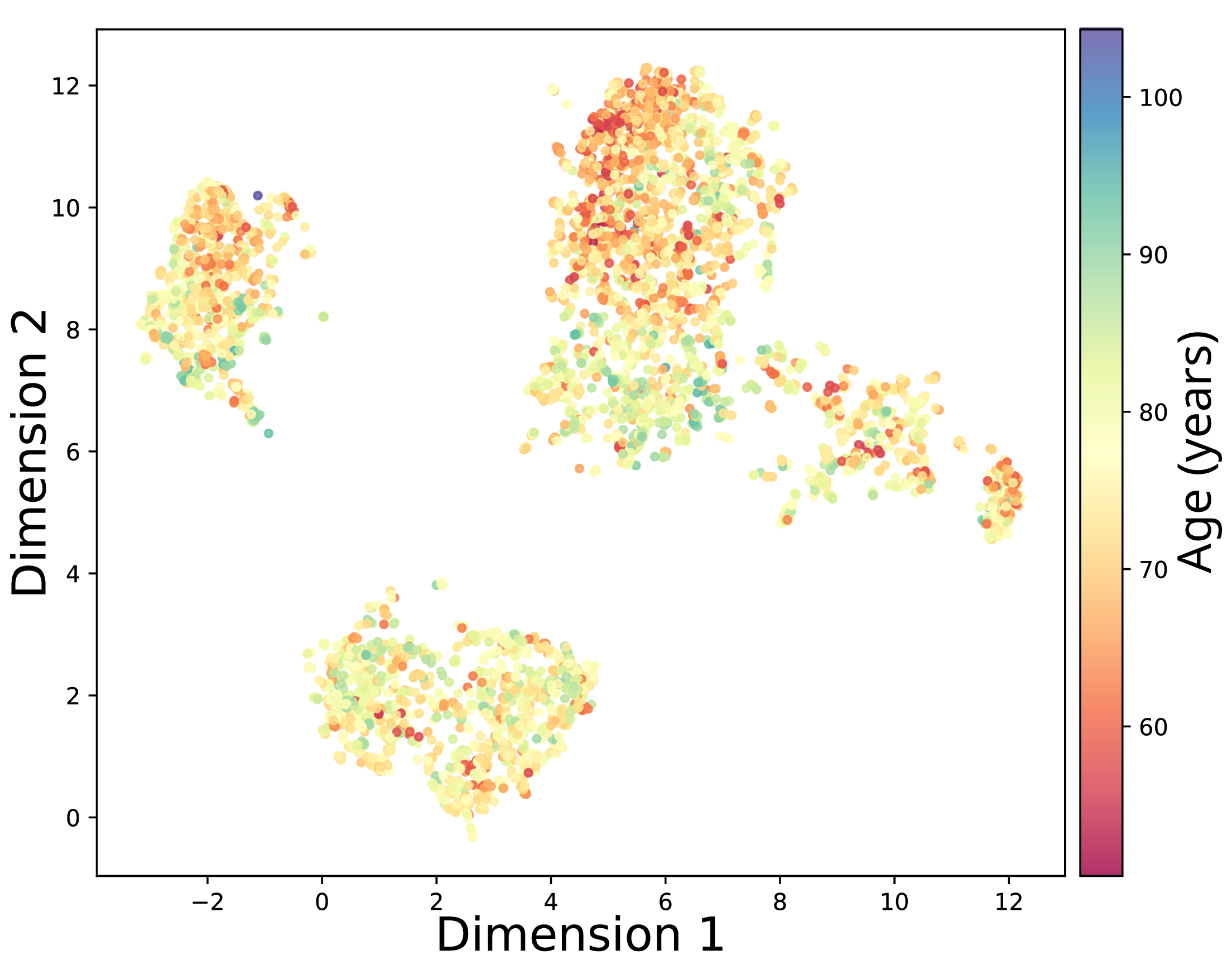}
         \label{fig:1a}
     \end{subfigure}
     \begin{subfigure}{0.49\textwidth}
         \centering
         \includegraphics[width=\linewidth]{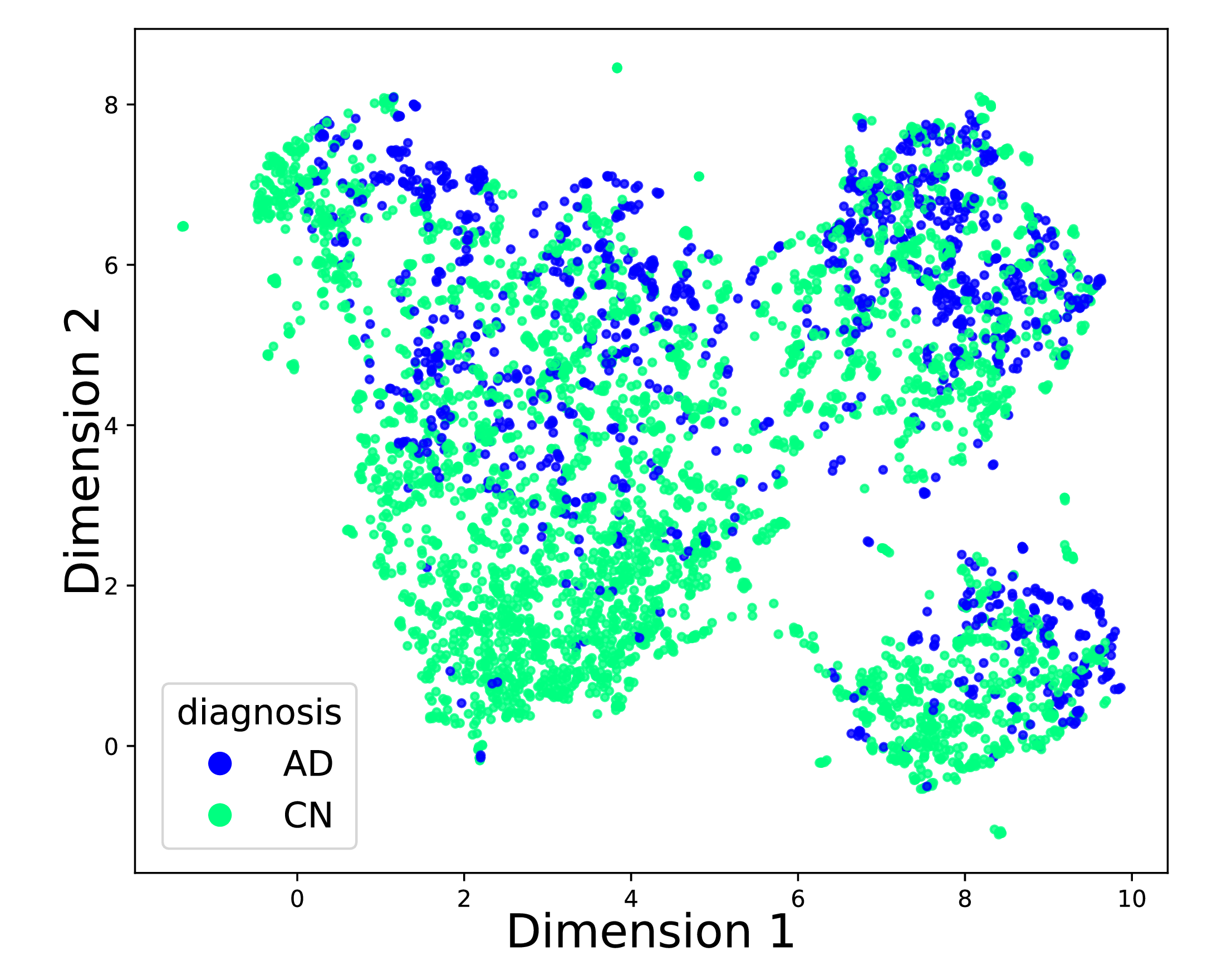}
         \label{fig:1b}
     \end{subfigure}
     \caption{UMAP projections of ADNI subject features: ResNet50 embeddings colored by age (left) and RadeImageNet features colored by diagnosis (right).}
     \label{fig:visu}
\end{figure}

Table \ref{tab:results} presents the results obtained for each model on all metrics of interest. 
The best model is the ViT pre-trained with DINOv2's method on the massive LVD-142M dataset, which achieves a median rank of 2 on the six metrics combined, ahead of ConvNext and ViT-DINO, which respective median ranks are 3 and 4. Among the 18 competing encoders, ViT-DINOv2 shows the best score on three out of the six considered tasks. 
Among models pre-trained on medical images, the best one is ResNet50 trained on RadImageNet, with a median rank of 5, showing the best performance on the AD diagnosis task and competitive performance on the other tasks like age regression. BiomedCLIP is the second-best medical model, ranking first on the PD diagnosis task, though the performance difference is not statistically significant. 
The six models pre-trained with SSL techniques all perform poorly in the benchmark occupying 5 of the last 6 places in the final ranking. 
Based on the statistical tests on models pairs, 
natural 2D encoders and RadImageNet 
tend to significantly outperform 
the other encoders (notably the 3D medical encoders), with no strong significant differences within either group.

Figure \ref{fig:visu} provides UMAP visualizations, with ages and diagnosis labels, of the ADNI latent space extracted from two encoders: ResNet50, the best-performing model for age regression, and RadImageNet, the best-performing model for AD diagnosis classification. The latent spaces shown demonstrate a tendency toward separability between AD and CN subjects (right figure), as well as slight age encoding (left figure).

\section{Discussions and conclusion}
In this work, we presented an extensive comparison of vision encoders for 3D brain MRI. Based on the performance on the different tasks shown in Table \ref{tab:results}, 2D encoders pre-trained on natural images (1st and 2nd series of rows) produce features of higher quality compared to both fully and self-supervised models pre-trained on medical images (3rd and 4th series). These findings align with previous works \cite{an2025raptor,gia2026uad_2D_foundation_zeroshot}. The same observations (data not shown) hold when using the Raptor method \cite{an2025raptor} to set the embedding dimensions from the 2D encoders at the same magnitude as the 3D encoders (approximately 512).

Noticeably, for both age regression and sex classification, the relative performances among the encoders are consistent across PPMI and ADNI. Age regression performances on ADNI control subjects are higher than those on PPMI ones, probably due to the substantial difference in the size of the healthy control subsets between the two datasets. Though, the sex classification performances are similar on both datasets. 
On the PD diagnosis task, accuracies achieved by the models are very low, ranging between 0.572 and 0.624 
with no significant differences between models. This is primarily due to the limited discriminative power of T1-weighted MRI for this task, as it does not captures functional or metabolic alterations (e.g. dopaminergic neuron loss in the substantia nigra) that are characteristic of PD. In contrast, the best balanced accuracy on the AD diagnosis task reaches $0.857$, 
with RadImageNet significantly outperforming most other encoders. T1-weighted MRI is highly relevant for AD, as it effectively captures structural atrophy patterns (i.e. volume loss) which serve as hallmark biomarkers for diagnosis.

Among the encoders trained on medical data, the one producing the most expressive latent space is the one trained on a classification task, RadImageNet-R50. 3D encoders trained with SSL fail to produce 
the most relevant latent representations, even while being trained on very large datasets: 114k volumes for the OpenMind's encoders, and 60K volumes for FOMO2JOMO's. \\


We have demonstrated that 2D encoders pre-trained on natural images consistently outperform those trained on medical datasets, even when the latter are trained on tens of thousands of 3D medical imaging volumes. This suggests that current 3D medical foundation models struggle to capture discriminative features as effectively as their 2D counterparts, highlighting a critical gap in representation learning for medical imaging.

Considering the benefits of recent breakthroughs in SSL for general computer vision and the increasing focus on medical applications, our findings underscore the need for more robust SSL strategies tailored to 3D medical data. We hope that 
our observations 
will motivate further research into representation learning methods for medical imaging. One possible way of improvement could be to include meta-data and clinical knowledge to help the learning process and compensate for the limited amount of data available for specific tasks \cite{zhang2023divessl}. Future work should explore the applicability of these representations to dense prediction tasks and their robustness across diverse imaging modalities and domains to fully validate their utility in medical image analysis.

\begin{credits}


\subsubsection{\ackname}
This work was partially funded by the French National Research Agency (ANR) through the
IMAGINA (ANR-18-CE17-0012), SEIZURE (ANR-24-CE45-4399), and BrainTwin (PEPR Santé Numérique 2026) projects.
This work was granted access to the HPC resources
of IDRIS under the allocation AD011017238 made by GENCI.

Data collection and sharing for the Alzheimer's Disease Neuroimaging Initiative (ADNI) is
funded by the National Institute on Aging (National Institutes of Health Grant
U19AG024904). The grantee organization is the Northern California Institute for Research
and Education. In the past, ADNI has also received funding from the National Institute of
Biomedical Imaging and Bioengineering, the Canadian Institutes of Health Research, and
private sector contributions through the Foundation for the National Institutes of Health
(FNIH) including generous contributions from the following: AbbVie, Alzheimer’s Association;
Alzheimer’s Drug Discovery Foundation; Araclon Biotech; BioClinica, Inc.; Biogen; Bristol-
Myers Squibb Company; CereSpir, Inc.; Cogstate; Eisai Inc.; Elan Pharmaceuticals, Inc.; Eli
Lilly and Company; EuroImmun; F. Hoffmann-La Roche Ltd and its affiliated company
Genentech, Inc.; Fujirebio; GE Healthcare; IXICO Ltd.; Janssen Alzheimer Immunotherapy
Research \& Development, LLC.; Johnson \& Johnson Pharmaceutical Research \&
Development LLC.; Lumosity; Lundbeck; Merck \& Co., Inc.; Meso Scale Diagnostics, LLC.;
NeuroRx Research; Neurotrack Technologies; Novartis Pharmaceuticals Corporation; Pfizer
Inc.; Piramal Imaging; Servier; Takeda Pharmaceutical Company; and Transition
Therapeutics.



Data used in the preparation of this article was obtained on 2026-02-18 from the Parkinson's
Progression Markers Initiative (PPMI) database
(\url{https://www.ppmi-info.org/access-data-specimens/download-data}),
RRID:SCR\_006431. For up-to-date information on the study, visit \url{https://www.ppmi-info.org}.
PPMI – a public-private partnership – is funded by the Michael J. Fox Foundation for Parkinson’s Research and funding partners, including AbbVie, Alamar Biosciences, Aligning Science Across Parkinson’s (ASAP), Arrowhead Pharma, Arvinas, AskBio, BIAL, BioArctic, Biohaven, BlueRock Therapeutics, Bristol Myers Squibb, Calico Labs, Capsida Biotherapeutics, Critical Path Institute, DaCapo Brainscience, Denali, Edmond J. Safra Foundation, Eli Lilly, Gain Therapeutics, GE Healthcare, Genentech, GSK, Insitro, Johnson \& Johnson Innovative Medicine, Lundbeck, Merck, Neumora, Neuron23, Novartis, Olink, Regeneron, Roche, Sanofi, Tenvie, UCB, Vanqua Bio, Voyager Therapeutics, The Weston Family Foundation.

%

\subsubsection{\discintname}
The authors have no competing interests to declare
that are relevant to the content of this article.
\end{credits}

%
%
%

\newpage 

\bibliographystyle{splncs04}
\bibliography{bibliography}

\end{document}